\documentclass[a4paper,twoside]{article}

\usepackage{epsfig}
\usepackage{subcaption}
\usepackage{calc}
\usepackage{amssymb}
\usepackage{amstext}
\usepackage{amsmath}
\usepackage{amsthm}
\usepackage{hyperref}
\usepackage{cleveref}
\usepackage{multicol}
\usepackage{booktabs}
\usepackage{multirow}
\usepackage{pslatex}
\usepackage{apalike}
\usepackage{algorithm, algpseudocode}
\usepackage[bottom]{footmisc}
\usepackage{SCITEPRESS}     

\newcommand{\startappendices}{%
  \setcounter{section}{0}%
  \renewcommand\thesection{\Alph{section}}%
}

\newcommand{\appsection}[1]{%
  \refstepcounter{section}
  \phantomsection
  \section*{#1}%
  \label{#1}%
}

\begin{document}

\title{Defense That Attacks: How Robust Models Become Better Attackers}


\author{\authorname{Mohamed Awad\sup{1, 2}, Mahmoud Mohamed\sup{2} and Walid Gomaa\sup{2, 3}}
\affiliation{\sup{1}Department of Statistics and Data Science, MBZUAI, Abu Dhabi, UAE}
\affiliation{\sup{2}Cyber-Physical Systems Lab,
	Egypt Japan University of Science and Technology,
	Alexandria, Egypt}
\affiliation{\sup{3}Faculty of Engineering,
	Alexandria University,
	Alexandria, Egypt}
\email{mohamed.awad@mbzuai.ac.ae , \{mahmoud.akrm, walid.gomaa\}@ejust.edu.eg}
}


\abstract{Deep learning has achieved great success in computer vision, but remains vulnerable to adversarial attacks. Adversarial training is the leading defense designed to improve model robustness. However, its effect on the transferability of attacks is underexplored. In this work, we ask whether adversarial training unintentionally increases the transferability of adversarial examples. To answer this, we trained a diverse zoo of \textbf{36 models}, including CNNs and ViTs, and conducted comprehensive transferability experiments. Our results reveal a clear paradox: adversarially trained (AT) models produce perturbations that transfer more effectively than those from standard models, which introduce a new ecosystem risk. To enable reproducibility and further study, we release all models, code, and experimental scripts. Furthermore, we argue that robustness evaluations should assess not only the resistance of a model to transferred attacks but also its propensity to produce transferable adversarial examples. The code and models can be found in the repo: \href{https://github.com/mohamed11354/Defense-That-Attacks-How-Robust-Models-Become-Better-Attackers}{https://github.com/mohamed11354/Defense-That-Attacks}.}

\onecolumn \maketitle \normalsize \setcounter{footnote}{0} \vfill

\section{Introduction}
\label{sec:introduction}

\par


Deep learning has transformed computer vision by achieving state-of-the-art results in image classification, object detection, and semantic segmentation. These advances driven by architectures such as convolutional neural networks (CNNs)~\cite{huang2017densely,he2016deep,tan2019efficientnet,howard2017mobilenets,simonyan2014very} and vision transformers (ViTs)~\cite{dosovitskiy2020image,liu2021swin,touvron2021training} have rapidly moved from research to real-world deployment in areas such as medical diagnosis, autonomous driving, and surveillance~\cite{li2023medical,zhang2023research}. As these systems are used in safety-critical environments, their reliability and security have become central concerns since failures can endanger lives or cause serious economic loss.

\begin{figure*}[thbp!]
	\centering
	\includegraphics[width=\linewidth]{heatmap_transfer_pgd_targets.png}\
	\caption{Heatmap of target model accuracy on our transferability benchmark. The X-axis lists surrogate models, with those trained using standard training on the left and adversarial training on the right. The Y-axis lists target models, all of which are adversarially trained. Colors represent target accuracy, with blue cells indicating lower accuracy and thus stronger transferability of adversarial examples.}
	\label{Heat map}
\end{figure*}

Despite their success, these models are highly vulnerable to adversarial examples~\cite{biggio2013evasion,goodfellow2014explaining,madry2017towards,carlini2017towards,ma2023transferable}. Adversarial examples are inputs formed by adding small perturbations constrained by a $\ell_p$-norm budget $\epsilon$, which cause the model to make incorrect predictions with high confidence. A key property of adversarial attacks is their transferability~\cite{papernot2016transferability}, where perturbations created from one model can also deceive other models, even when they differ in architecture or training data. This transferability enables black-box attacks. For example, an attacker can train a surrogate model, which is a model with accessible parameters and weights, to generate adversarial examples that are then applied to a protected target model whose parameters are unknown~\cite{gu2023survey}. As deep learning systems continue to spread to critical domains, these vulnerabilities are no longer theoretical but represent practical security risks that demand strong and reliable defenses.

In response to these security threats, the research community has developed numerous defense mechanisms aimed at mitigating adversarial vulnerabilities. Early approaches often relied on gradient obfuscation techniques that attempted to hide model gradients from attackers. These defenses were later shown to provide a false sense of security, as they can be easily circumvented by adaptive attacks rather than truly strengthening the model's robustness.\cite{athalye2018obfuscated,zhang2024gradient}. In contrast, adversarial training has emerged as the most reliable defense, where models are explicitly trained on adversarially perturbed data to withstand worst case attacks \cite{madry2017towards}. Building on this foundation, several variants such as TRADES and iGAT have emerged to improve its efficiency and the trade off between robustness and accuracy \cite{zhang2019theoreticallyprincipledtradeoffrobustness,deng2023understandingimprovingensembleadversarial}.

Although the transferability of adversarial examples between standard models has been extensively studied~\cite{su2018robustness,shao2021adversarial,bai2021transformers}, the impact of adversarial training on this phenomenon remains underexplored. Because adversarial training reshapes feature representations and gradient landscapes, it may alter how perturbations propagate across architectures. This raises a critical question: Could a defense designed to enhance robustness inadvertently make AT models stronger surrogate attackers? If so, it would create an ecosystem-level risk, where strengthening the defenses of one model might weaken others against black-box attacks.

To examine this possibility, we constructed a diverse model zoo containing 36 architectures, including both CNNs and VITs, all trained with identical adversarial training protocols to ensure that our findings reflect the true impact of adversarial training on transferability. Contrary to intuitive expectations that robust models should produce less transferable attacks, our analysis reveals a consistent pattern. AT models generate adversarial examples that transfer significantly more effectively than those from standard trained (ST) counterparts. This outcome indicates that robust features acquired through adversarial training may unintentionally introduce universal vulnerabilities that extend across architectures, as summarized in~\Cref{Heat map}. Our key contributions are as follows.

\begin{itemize}
    \item We conducted large-scale adversarial training on both CNN and ViT architectures, covering 36 models in total. All trained models, along with the training code, will be publicly released in a GitHub repository to support reproducibility and further research.  
    \item We introduced a benchmark for adversarial example transferability using MIG with $\epsilon = 16$, showing that AT models consistently produce more transferable attacks.  
    \item We establish that AT models generate more transferable attacks, revealing a security paradox that challenges current assumptions about robustness.  
\end{itemize}

\begin{figure*}[thbp!]
	\centering
	\begin{multicols}{2}
		\includegraphics[width=\linewidth]{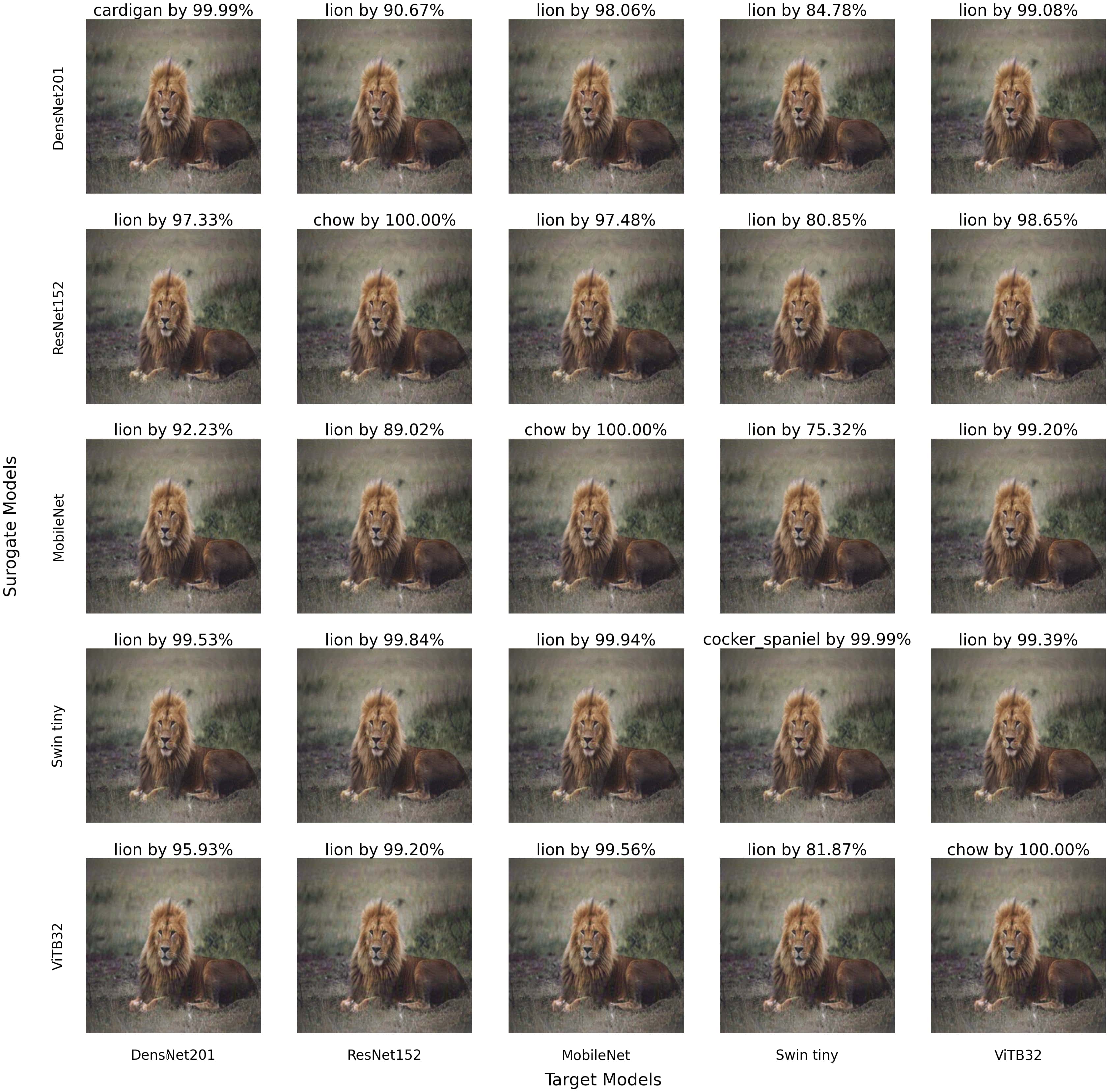}\par 
		\includegraphics[width=\linewidth]{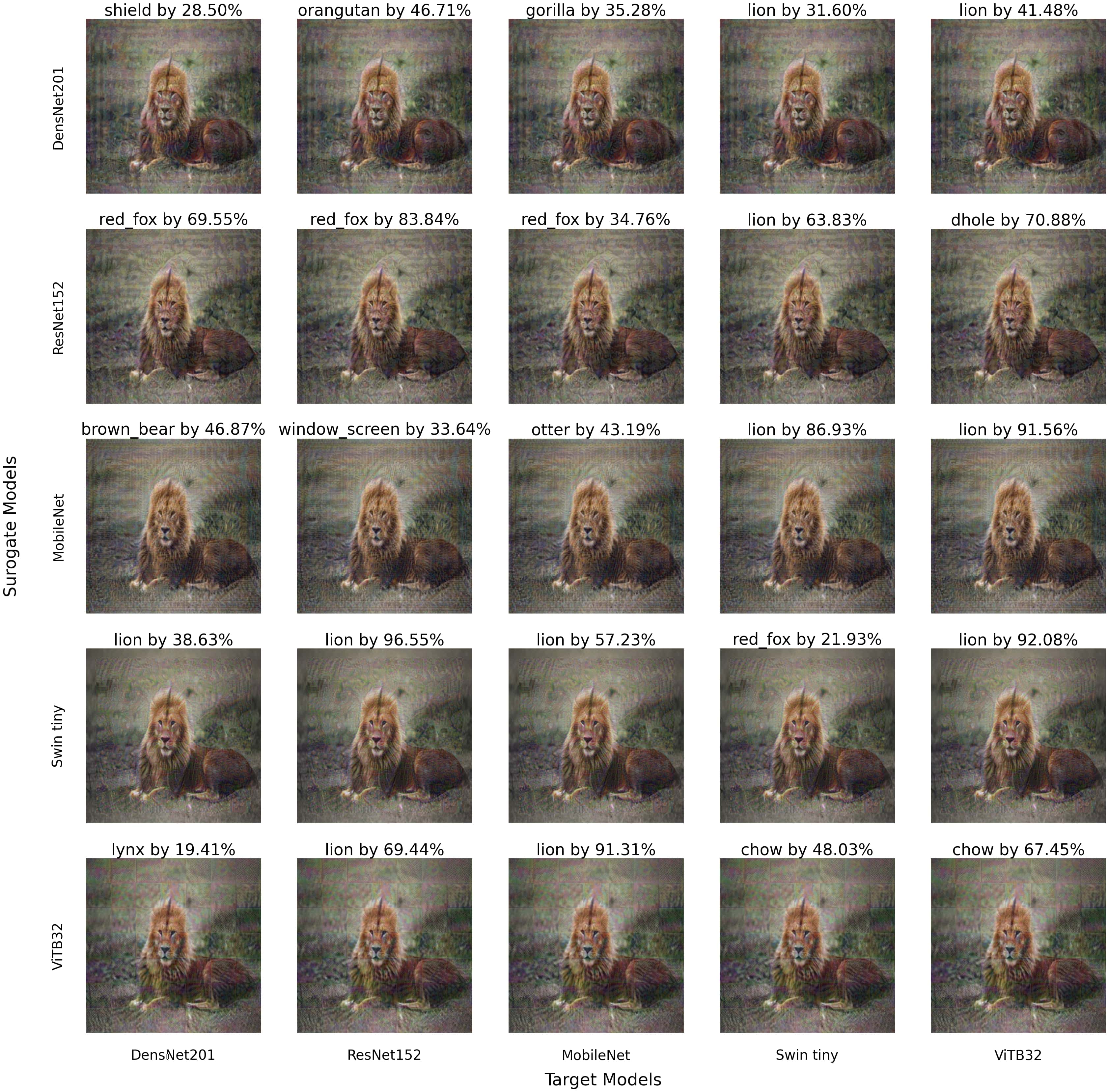}\par 
	\end{multicols}
	\caption{The transferability of PGD (left image) and MIG (right image) across five different models with attacks generated using $\epsilon = 16$.
	}
	\label{Figure 3.}
\end{figure*}

\section{Background}
\label{sec:background}
In this paper, we focus on \textbf{untargeted $\ell_\infty$-bounded adversarial attacks} with perturbation budget~$\epsilon$. 
Let an input image with $d$ channels be denoted by $x \in [0,255]^d$ and its ground-truth label by $y$. 
We consider a classifier $f_\theta$, optimized using a loss function~$\mathcal{L}$, and an attack step size~$\alpha$. 
The perturbation space is defined as the $\ell_\infty$ ball centered at $x$ with radius~$\epsilon$, namely
\[
B_\infty(x, \epsilon) = \{ \tilde{x} : \| \tilde{x} - x \|_\infty \leq \epsilon \}.
\]
We use $\Pi_{B_\infty(x, \epsilon)}$ to denote the projection operator (i.e., clipping) on this set and $\delta = \tilde{x} - x$ to represent the change between a clean and adversarial image.

\subsection{Attacks}
\paragraph{Projected Gradient Descent} PGD is an iterative attack that starts with clean input \(x\) and does \(T\) iteration updates according to maximizes the loss \(\mathcal{L}(f_\theta(x),y)\) to obtain an adversarial image \(x_T\). The updates follow the rule
\begin{equation}
x_{t+1}
=\Pi_{B_\infty(x,\epsilon)}\!\Big(
x_t+\alpha\,\operatorname{sign}\big(\nabla_{x_t}\mathcal{L}(f_\theta(x_t),y)\big)
\Big)
\label{eq:pgd}
\end{equation}
Here, \(\alpha\) is a step size that controls the update in each iteration; generally, it is set to \(\epsilon/T\). For sufficiently small \(\epsilon\), \(x_T\) is visually close to \(x\) while increasing the loss; larger \(\epsilon\) typically improves attack success but reduces perceptual similarity.

\paragraph{Momentum Integrated Gradient}MIG~\cite{ma2023transferable} is a promising attack that is implemented upon integrated gradient (IG)~\cite{sundararajan2017axiomatic}. MIG updates perturbation iteratively through a momentum strategy directed by IG. The attack focuses on manipulating the most influential components identified by the IG, increasing the likelihood of misclassifications. A critical feature of IG, known as \textbf{implementation invariance}. This feature allows attacks crafted by MIG to be effectively transferred between different architectures, such as CNN and ViT, or even between both types of models. This transferability power did not exist on previous attacks such as PGD, as shown in \Cref{Figure 3.}.

IG is an axiomatic interpretability method designed to identify the most influential features contributing to a model’s predictions. It is computed by accumulating gradients along a straight-line path from a baseline input to the actual input. The baseline represents a reference point in the input space that contains no meaningful information.

In image classification tasks, the baseline is typically chosen as a completely black image. The path from the baseline to the input image can be represented as a sequence of intermediate images with gradually increasing brightness. Formally, this corresponds to scaling the input image $x$ by a series of values $\in [0,1]$. In our setting, we fix the baseline to black image. 
The Integrated Gradients (IG) attribution for a single pixel $i$ in an input image $x$, with 
baseline $b$, for a model $f$, is approximated as follows:

\begin{equation}
	IG_i(f,x,b) \approx (x_i-b_i)\times \sum_{k=1}^{s}\frac{\partial f( b + \frac{k}{s} \times (x-b)) }{\partial x_i} \times \frac{1}{s}
\end{equation}

Here, $s$ denotes the approximation step, which represents the number of intermediate brightness levels used for an input image. In other words, $s$ corresponds to the number of steps taken to approximate the integration, which controls the trade-off between computational cost and approximation accuracy. In our experiments, we fixed $s = 20$ to achieve a reasonable balance between efficiency and accuracy.

MIG attack adds a parameter $\mu$ that serves as the momentum factor that regulates the decay rate of old gradients. In the original work~\cite{ma2023transferable}, the momentum factor was demonstrated to achieve optimal attack success rates at $\mu = 1$, where deviations above or below this value reduce the attack's effectiveness. 

\subsection{Adversarial training}

Standard training aims to learn parameters \(\theta\) by minimizing the expected loss \(\mathbb{E}_{(x,y)\sim\mathcal{D}}[\mathcal{L}(f_\theta(x),y)]\). Adversarial training extends the standard training by minimizing the worst-case loss over norm-bounded perturbations around each input. Throughout this paper, we adopt the \(\ell_\infty\) threat model with budget \(\epsilon\), leading to the min--max objective of~\cite{madry2017towards}:
\begin{equation}
\label{eq:madry_at}
\min_{\theta}\;
\mathbb{E}_{(x,y)\sim\mathcal{D}}
\Big[
\max_{\|\delta\|_\infty \le \epsilon}\,
\mathcal{L}\!\big(f_\theta(x+\delta),\,y\big)
\Big].
\end{equation}
The inner maximization is typically approximated by projected gradient methods (e.g., multistep PGD), while the outer minimization is updated \(\theta\) using stochastic gradients.

In our work, we adopt \emph{Free Adversarial Training}~\cite{shafahi2019adversarial} as a variant implementation of equation ~\eqref{eq:madry_at}, which reduces the overhead of PGD-based Adversarial Training by replaying each mini-batch \(K\) times and sharing a single backward pass to update both the model parameters and adversarial perturbations.
Concretely, for a mini-batch \((x,y)\) with a persistent perturbation \(\delta\in[-\epsilon,\epsilon]^d\), each replay performs
\begin{align}
g_x &\leftarrow \nabla_x\,\mathcal{L}\!\big(f_\theta(x+\delta),\,y\big), 
\quad
g_\theta \leftarrow \nabla_\theta\,\mathcal{L}\!\big(f_\theta(x+\delta),\,y\big), \\
\tilde{x} &\leftarrow \Pi_{B_\infty(x,\epsilon)}\!\Big(x+\delta+\alpha\,\mathrm{sign}(g_x)\Big), \\
\delta &\leftarrow \tilde{x} - x, 
\quad 
\theta \leftarrow \theta - \eta\, g_\theta.
\end{align}
where \(\eta\) is the learning rate. Compared to \(K\)-step PGD-AT, this procedure amortizes the inner-loop cost by reusing the same backpropagation for both updates and moving the \(K\) adversarial steps into the training loop via mini-batch replays. In this paper, ``AT model'' refers to a model trained using free adverserial training algorithm under the \(\ell_\infty\) norm.

\subsection{Threat model}
We define a \emph{surrogate} (or \emph{source}) model as any model whose parameters are accessible to the attacker and on which adversarial examples are generated. A \emph{target} model is the intended victim model whose parameters are not accessible. \emph{Transferability} refers to the phenomenon that adversarial examples crafted on a surrogate also cause misclassification on a distinct target.

We study \emph{transfer-based black-box} attacks under an $\ell_\infty$ threat model. In this setting, the attacker has full access to one or more surrogate models (including their architectures, parameters, and gradients) and generates adversarial examples using attacks such as PGD or MIG. Crucially, the attacker has \emph{no access} to the target model’s parameters or gradients and issues \emph{zero queries} to the target during attack generation. This contrasts with \emph{white-box} attacks, where the attacker has complete access to the target model, and with \emph{query-based black-box} attacks, where the attacker can adaptively query the target to guide the generation of adversarial examples.

\section{Methodology}
\label{sec:method}

\subsection{Model zoo construction}
\label{sec:model_zoo}
We constructed a comprehensive model zoo of 36 diverse architectures spanning major convolutional families and vision transformers. The zoo includes representative models from DenseNet~\cite{huang2017densely}, MobileNet~\cite{howard2017mobilenets,sandler2018mobilenetv2}, ResNet~\cite{he2016deep,bello2021revisiting}, EfficientNet~\cite{tan2019efficientnet,tan2021efficientnetv2}, NASNet~\cite{zoph2018learning}, and Vision Transformers~\cite{dosovitskiy2020image}.

For the standard training condition, we utilized pre-trained weights from TensorFlow's model zoo. For adversarial training we applied Free Adversarial Training~\cite{shafahi2019adversarial} to the same architectures to ensure comparability between ST and AT models. The Training hyperparameters are provided in full in Appendix~\ref{training recipe}.

\subsection{Robustness evaluation of the constructed zoo}
\label{sec:robust_eval}
Model robustness was evaluated using three approaches: clean accuracy ($\varepsilon=0$), PGD-20, and MIG attacks. Epsilon values are expressed as integer pixel magnitudes in $[0,255]$; for instance, $\epsilon=8$ corresponds to a perturbation budget of $8/255$.  Exact attack parameters for all methods are listed in Appendix~\Cref{complete tables}.

To reduce computational cost during model evaluation, we modified the standard MIG implementation by generating perturbations at $\epsilon=5$ and clipping for smaller epsilon values. This approach reduced computational overhead while maintaining attack effectiveness, as validated in Appendix~\Cref{Mig mod}. PGD-20 evaluations used 50,000 validation images, while MIG attacks used 1,600 randomly selected images.

\subsection{Controlled transferability analysis design}
\label{sec:transfer_design}

In principle, our model zoo spans 36 architectures, each available in both ST and AT modes. 
This results in $72$ distinct models in total. 
Accordingly, the complete transferability space would involve $72 \times 72$ surrogate--target pairs. Exhaustively evaluating all of them would be computationally prohibitive and difficult to interpret. Instead, we designed a structured subset of experiments to focus on the main factors of interest: architecture, training type, and model capacity.  

For this subset, we chose eight surrogates and sixteen targets. The surrogates consist of smaller variants of four representative families, which are DenseNet121, EfficientNet-B0, MobileNetV2 and ViT-Tiny/16. Each of them was trained in both ST and AT settings. The targets are larger models from diverse families, including DenseNet, ResNet, ResNetRS, EfficientNet, EfficientNetV2, MobileNetV3-Large, NASNetLarge and ViT-Base/16, also trained in ST and AT modes.  

This design allows for the isolation of three dimensions of transferability. First, architecture-specific effects: whether transferability is stronger within the same family, and how CNN$\to$ViT and ViT$\to$CNN transfers behave differently. Second, training-type interactions: whether AT$\to$ST attacks transfer differently from ST$\to$AT, and how transferability changes under homogeneous conditions (AT$\to$AT, ST$\to$ST). Third, capacity and attack-strength effects: whether adversarial examples crafted on small surrogates transfer to large targets, and how this relationship evolves across different perturbation budgets.  

\subsection{Transferability benchmark and evaluation metric}
\label{sec:benchmark_metric}
We build a transferability benchmark by generating adversarial examples on each surrogate model using unmodified MIG on a high budget $\varepsilon=16$. The benchmark set contains 2,000 clean images sampled uniformly from ImageNet validation (two images per class) and the corresponding adversarial variants generated on each surrogate model.

We measure the \emph{target accuracy on adversarial examples} produced by a surrogate, which directly shows the vulnerability of a target model to transferred attacks. Higher transferability corresponds to lower accuracy across models.

\section{Results}
\subsection{Training-Type Interaction: The Adversarial Training Paradox}
\label{sec:results_training_type}

 We summarized the result of this experiment in Table~\ref{tab:training_type}; each cell is the average target accuracy on adversarial examples from a set of surrogates and targets (averages are computed over 4 surrogates by 8 targets, i.e., 32 surrogate--target pairs, evaluated on the 2,000-image benchmark described in Section~\ref{sec:method}).

To avoid ambiguity, we clarify the notation used throughout this section. ``$A \to B$'' means that adversarial examples were generated from the surrogate set $A$ and then applied to the target set $B$. For example, ``AT \(\to\) ST'' denotes adversarial examples that were crafted from AT surrogates and applied to a ST target.


\begin{table}[ht]
\centering
\caption{Training-type transferability matrix (averages over $4\times 8$ pairs). Lower target accuracy means stronger transfer.}
\label{tab:training_type}
\begin{tabular}{lcc}
\toprule
 & \multicolumn{2}{c}{\textbf{Target}} \\
\cmidrule(lr){2-3}
\textbf{Surrogate} & \textbf{AT} & \textbf{ST} \\
\midrule
\textbf{AT} & 13.16\% & 26.58\% \\
\textbf{ST} & 48.09\% & 32.88\% \\
\bottomrule
\end{tabular}
\end{table}

The most striking outcome of our experiments appears in the scenario where AT models are attacked by adversarial examples crafted by other AT models. In this case the target accuracy falls to 13.16$\%$ on average. This result is notable because it represents a large improvement in attack effectiveness relative to the ST $\to$ ST baseline, where target accuracy is 32.88$\%$. Rather than simply improving model security, AT appears to produce models that are highly effective at attacking one another. The AT $\to$ AT condition is even stronger than the AT $\to$ ST case where target accuracy is 26.58$\%$, which suggests that AT induces similar robust features across models and thus a shared vulnerability.

The results reveal a clear ordering of transferability across training types. AT sources dominate standard sources since their adversarial examples reduce target accuracy more across the board. Standard sources are least effective against robust targets, as shown by the ST $\to$ AT condition with 48.09$\%$ accuracy. In other words robust defenses block attacks from standard surrogates but are surprisingly fragile to attacks from fellow robust models.

These results show that while AT increases robustness under direct attacks, it also changes the learned representations such that adversarial perturbations become more effective when transferred to other models. This creates an ecosystem level concern. Widespread use of AT could unintentionally produce a population of surrogate models that are especially powerful for transfer based black-box attacks, shifting the security problem from single model hardening to system level vulnerability. Full per-pair results are provided in Appendix~\ref{complete tables}.

\begin{figure}[thbp!]
	\centering
	\includegraphics[width=\linewidth]{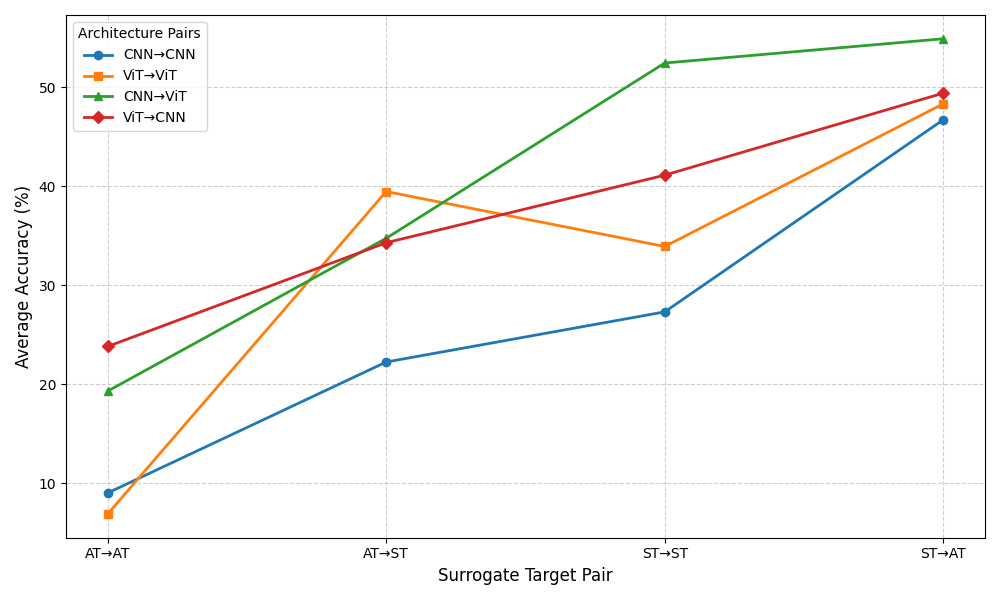}\
	\caption{
    The architectural trend across different surrogate target pairs.  
	}
	\label{lineplot}
\end{figure}





\begin{figure*}
    \centering
     \includegraphics[width=1.75\columnwidth]{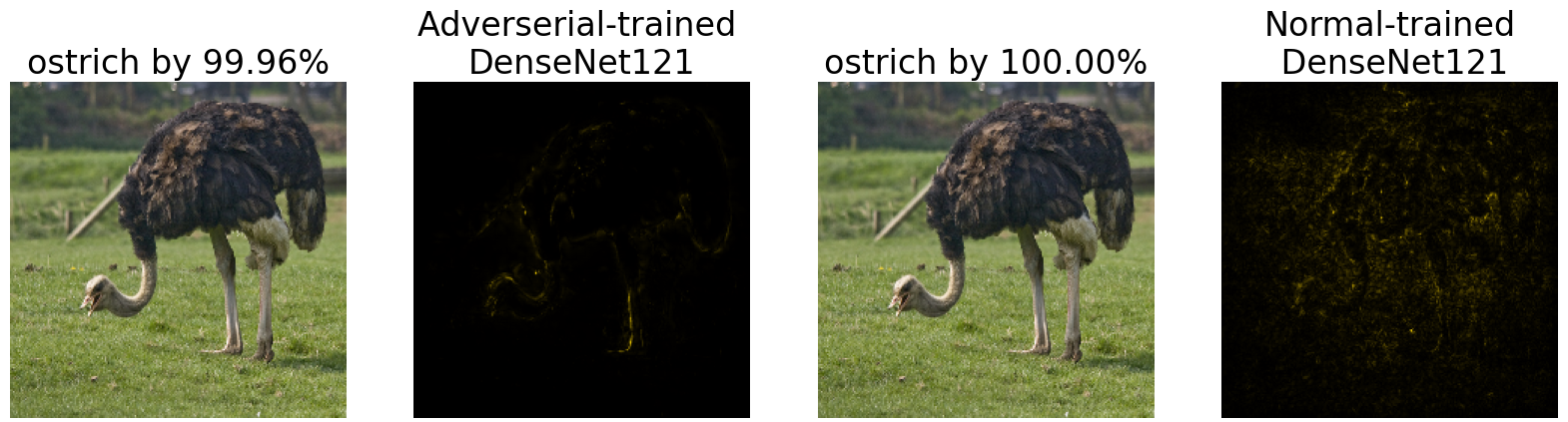}
    \includegraphics[width=1.75\columnwidth]{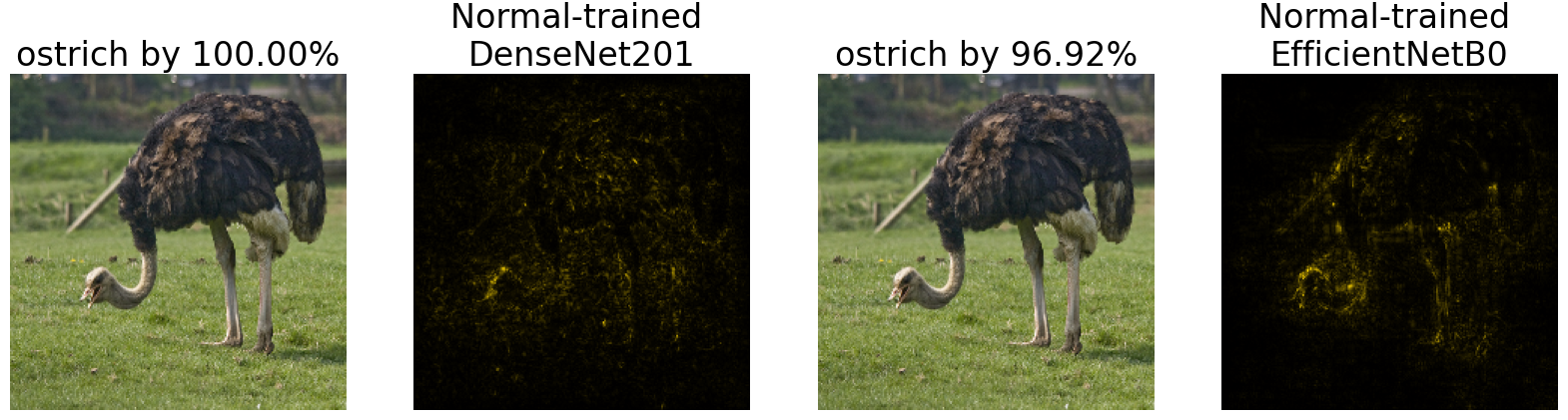}
	\caption{Integrated gradients (IG) on both AT and ST models}
	\label{Ostrich}
\end{figure*}
\subsection{Architecture-Specific Transferability Patterns}
\label{sec:architecture_patterns}

Building on the training-type paradox established previously, we now examine how this phenomenon behaves across different architectural relationships. The key question is whether the strong AT$\to$AT transferability observed earlier arises primarily from the training methodology itself or from the architectural choices of the models involved. Figure~\ref{lineplot} presents a comprehensive view of transferability patterns across four key scenarios: CNN$\to$CNN, ViT$\to$ViT, CNN$\to$ViT, and ViT$\to$CNN transfers.

As clearly visualized in the line graph (Figure~\ref{lineplot}), AT surrogates consistently generate more potent transferable attacks than their ST counterparts, indicating that this phenomenon is architecture-agnostic. This is evident from the fact that the first two data points in each series, which represent scenarios where an \textit{AT model is the surrogate}, consistently result in lower target model accuracy than the subsequent two points, which represent \textit{ST model surrogates}. The only minor deviation from a perfectly smooth curve occurs in the ViT$\to$ViT case, a phenomenon we attribute to the unique feature space similarity between models and which is analyzed in detail in the following section.

The strongest effect appears in the AT$\to$AT scenario, where AT models attack other robust models. Here, CNN$\to$CNN transfers achieve remarkably low target accuracy of 9.02\%. The ViT$\to$ViT scenario shows similar vulnerability with 11.95\% accuracy, Together, these results indicate that both model families inherit the same fundamental weakness when AT models attack one another.

When examining cross-architecture transfers, the same pattern emerges. Attacks by AT surrogates achieve lower accuracies than those from ST surrogates, both in CNN$\to$ViT and ViT$\to$CNN cases. which indicates that the increased transferability induced by AT is not limited to models of the same family but also extends across different architectures.

Taken together, these results show that while the baseline level of transferability varies across architectures, the relative advantage of AT surrogates is stable. This suggests that the changes induced by AT produce attack vectors that generalize across both similar and dissimilar architectures. The adversarial training transferability paradox is not an artifact of specific architectural choices but rather a fundamental property of how adversarial training reshapes model feature spaces. The consistency of this effect across diverse architectural relationships underscores its significance for ecosystem-level security considerations.

\subsection{Understanding the Mechanism}
\label{sec:qualitative}

To investigate why AT increases transferability we inspected model attributions using integrated gradients. Figure~\ref{Ostrich} shows representative attributions for AT and ST variants of DenseNet121 and illustrates a clear difference in how these models represent images.

AT models tend to produce more semantically oriented attributions. For the AT DenseNet121 the integrated gradients highlight regions such as the head, body and legs of the ostrich instead of only textural details. These specific, object-level attributions indicate that AT models rely on features that capture high level visual concepts rather than narrowly focusing on local textures or background cues.

Standardly trained models show the opposite pattern. Their attributions are distributed over local texture and fine detail. In the DenseNet example the ST model places strong weight on feather textures and background patches. Such localized focus makes ST perturbations more tight to a model and less likely to transfer across different architectures.

This difference in attribution explains the transferability effect. Perturbations that disrupt semantically meaningful features are more likely to affect many models because those features reflect common visual concepts. When an AT surrogate crafts an adversarial example it tends to target these general features, producing perturbations that remain effective against diverse targets. That mechanism accounts for the consistently higher transferability of AT sources observed.

The analysis also clarifies why AT models are vulnerable to attacks from other AT models. Different AT models learn to focus on the same semantically meaningful regions, which creates a shared vulnerability. An adversarial example that successfully attacks those regions will transfer effectively between AT models, which explains the low AT$\to$AT accuracies reported earlier. While this effect is most pronounced for robust models it is not unique to them. Similar vulnerabilities can also emerge in ST models when the source and target architectures are closely related. 

There are notable exceptions that highlight the role of representation similarity. Pairs of closely related architectures can show strong transfer even in ST form. Examples in our benchmark include DenseNet121 $\to$ DenseNet201, EfficientNet-B0 $\to$ EfficientNet-B7, and ViT-Tiny $\to$ ViT-Base. In these cases the surrogate and target learn highly similar feature extractors so even localized ST perturbations can transfer effectively. Figure~\ref{Ostrich} illustrates this for DenseNet121 and DenseNet201 where integrated gradients reveal visually similar and overlapping representations. MobileNetV3 behaves similarly because its newer design incorporates elements from EfficientNet, which increases its similarity to those models and its vulnerability to EfficientNet surrogates. See~\Cref{transfer table2} for the highlighted pairs.

EfficientNet-B0 also provides an illustrative special case. Even without AT it exhibits more distributed attributions than many other ST models. This architectural tendency toward broader feature use helps explain why EfficientNet-B0 is often an effective surrogate in ~\Cref{transfer table2}; its perturbations already target relatively general features.

Despite these same-architecture examples, architectural similarity is not a dependable assumption in real world black box scenarios where the target architecture is unknown. In such settings attackers cannot rely on finding a closely related surrogate and instead prefer surrogates whose perturbations transfer broadly. Our results show that AT surrogates produce perturbations that are effective across diverse families, so AT models offer a practical advantage as surrogate choices when the target is unknown.

In short, integrated gradients tie the empirical findings to feature learning. AT changes what models learn. It shifts attention from narrow, model related cues toward descriptive, semantically meaningful features. That shift improves white-box robustness but also creates more universally exploitable directions in input space. The current AT approach, while effective for individual model defense, may inadvertently create ecosystem-level vulnerabilities by homogenizing feature representations across models. Future defenses should aim to preserve per-model robustness without creating such common attack surfaces.

\section{Conclusion}
\label{sec:conclusion}
\vspace{-5pt}
We presented a large empirical study showing a surprising trade off in adversarial robustness. Using a diverse model zoo, we find that AT models produce adversarial examples that transfer more effectively than standard models, with the strongest transfers occurring when robust surrogates attack robust targets (AT $\to$ AT). Our analyses suggest a common cause: adversarial training shifts representations toward more semantically meaningful features that different models share, and perturbations that attacks those features transfer across architectures. The practical consequence is that improving a single model's white box robustness can increase ecosystem vulnerability to transfer-based attacks. We release all models and code to support follow up work.

\vspace{-15pt}

\section{Future work}
Our paper opens several promising directions building on our dataset and published models. 
\begin{itemize}
    \item Extend the analysis to a broader family of vision backbones beyond the foundational transformer used here, including Data-Efficient Image Transformer (DeiT), Swin transformers, hybrid CNN-Transformer architectures, and recent Mamba-based vision models. 
    \item Explore alternative adversarial training objectives that replace PGD in inner maximization with integrated gradients , and rigorously evaluate their impact on robustness and black-box transfer success across these newer backbones.
\end{itemize}

\vspace{-15pt}
\section*{ACKNOWLEDGMENT}
This manuscript was edited with the assistance of artificial intelligence systems. In particular, we used OpenAI GPT-5 and DeepSeekV3 models to help refine wording for all sections of the paper. All AI-generated text was reviewed and corrected by the authors. This Research supported with Cloud TPUs from Google’s TPU Research Cloud (TRC). This work is Funded by the Science and Technology Development Fund STDF (Egypt); Project id: 51399 - ``VERAS: Virtual Exercise Recognition and Assessment System''.

\vspace{-15pt}
\bibliographystyle{apalike}
{\small
\bibliography{main}}

\startappendices
\vspace{-20pt}
\appsection{Adverserial Training Hyperparameters} %
\label{training recipe} 
\par 
Our experiments utilize the TPU Research Cloud (TRC) program, all hyperparameters are listed in~\Cref{hyperparameters}. For the batch size and learning rate scheduler parameters, adjustments are dependent on the number of TPU replicas. For TPU v2 and v3, there are 8 replicas, while for TPU v4, there are 4 replicas, thus requiring appropriate modifications. Regarding decay steps, 5004 represents the number of batches per epoch, and the learning rate decays every 8 epoches. The replay parameter ($m$) is necessary because the step is computed when the optimizer step function is called, which happens $m$ times per batch. 

\par
For the CNN models, we exclusively relied on TensorFlow's preprocessing functions; no additional augmentation or preprocessing techniques were applied. We also used an early stopping mechanism to dynamically determine the optimal number of epochs based on the validation loss during clean (non-adversarial) validation. While the model was trained on adversarial images, validation was performed on clean images. 

\par 
For ViTs, applying the same techniques as CNNs leads to accuracy degradation due to the incompatibility of stochastic gradient descent (SGD) with transformers \cite{liu2020understanding}. Instead, we used the approach outlined in~\cite{steiner2021train}. The ViT architecture itself was implemented using Keras-CV, and for preprocessing, we applied the inception-crop method. For training the ViT models, we utilized a fixed number of epochs $300$.

\begin{table}[th!]
	\centering
    \tiny
	\caption{Hyperparameters used in training.}
	\label{hyperparameters}
	\begin{tabular}{c|c|c}
		
		\multicolumn{1}{l|}{(pre-) training config}           & \multicolumn{1}{l|}{CNNs}                         & VITs                        \\ \hline
		\multicolumn{1}{l|}{batch size}                       & \multicolumn{1}{l|}{32*8}    & 32*8\\
		\multicolumn{1}{l|}{image size}                       & \multicolumn{1}{l|}{224x224}                      & 224x224                    \\
		\multicolumn{1}{l|}{classifier\_activation}           & \multicolumn{1}{l|}{softmax}                      & softmax                    \\ \hline
		\multicolumn{3}{c}{Optimizer}                                                                                                          \\ \hline
		\multicolumn{1}{l|}{optimizer}                        & \multicolumn{1}{l|}{SGD}                          & Adam                       \\
		\multicolumn{1}{l|}{momentum}                         & \multicolumn{1}{l|}{0.9}                          & N/A                        \\
		\multicolumn{1}{l|}{weight decay}                    & \multicolumn{1}{l|}{0.0001}                       & 0.1                        \\
		\multicolumn{1}{l|}{global clipnorm}                 & \multicolumn{1}{l|}{None}                         & 1.0                        \\ \hline
		\multicolumn{3}{c}{Learning Rate}                                                                                                      \\ \hline
		\multicolumn{1}{l|}{scheduele decay}                   & \multicolumn{1}{l|}{ExponentialDecay}             & CosineDecay                \\
		\multicolumn{1}{l|}{initial learning rate}          & \multicolumn{1}{l|}{0.1}                          & 0.001                      \\
		\multicolumn{1}{l|}{decay steps}                     & \multicolumn{1}{l|}{8*5004 *m} & 8*270\\
		\multicolumn{1}{l|}{decay rate}                      & \multicolumn{1}{l|}{0.1}                          & N/A                        \\
		\multicolumn{1}{l|}{staircase}                        & \multicolumn{1}{l|}{True}                         & N/A                        \\
		\multicolumn{1}{l|}{warmup target}                   & \multicolumn{1}{l|}{N/A}                          & 0.001                      \\
		\multicolumn{1}{l|}{warmup steps}                    & \multicolumn{1}{l|}{N/A}                          & 8*30\\ \hline
		\multicolumn{3}{c}{Adverserial Attack}                                                                                                 \\ \hline
		\multicolumn{1}{l|}{replay parameter (m)}             & \multicolumn{1}{l|}{4}                            & 4                          \\
		\multicolumn{1}{l|}{step size ($\alpha$)}  & \multicolumn{1}{l|}{0.6}                          & 0.6                        \\
		\multicolumn{1}{l|}{perturbation budget ($\epsilon$)} & \multicolumn{1}{l|}{2}                            & 2                         
	\end{tabular}
\end{table}

\begin{figure}[htbp!]
	\centering
	\includegraphics[width=\columnwidth]{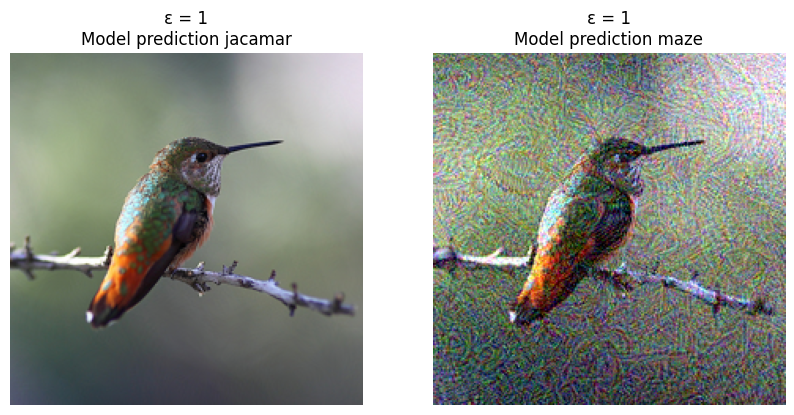}
	\caption{PGD-20 for hummingbird image on two MobileNet implementations, left image is the model with internal preprocessing layer while the right uses preprocessing externally. }
	\label{Figure 4.}
\end{figure}

\par
In TensorFlow~\cite{tensorflow2015-whitepaper}, the implementation of CNNs varies, particularly regarding the inclusion of the preprocessing layer within the model or keeping it external. For example, MobileNet has an external preprocessing function that rescales image pixels from $[0,255]$ to $[-1,1]$ meaning an epsilon value of $1$ would have an impact equivalent to  $\epsilon = 128$ if preprocessing were done internally. \Cref{Figure 4.} illustrates PGD conducted on MobileNet, comparing instances with preprocessing embedded within the model architecture to those with preprocessing conducted externally. 
Comparing the impact across different model families posed a challenge due to variations in the preprocessing procedures. To ensure consistency, we inserted a preprocessing layer into all models lacking one.

		
		
	

\vspace{-20pt}

\appsection{MIG modification} %
\label{Mig mod}

\vspace{-7pt}
\par
The MIG attack is slow to craft due to the approximation steps required to calculate the integrated gradient, which creates a bottleneck by converting a single image into a batch of scaled brightness images. Based on our empirical work, the MIG attack takes approximately seven times longer than the PGD-20 attack. Consequently, our evaluation of PGD-20 with five different epsilon values took less time than a single MIG run, despite the optimizations made. The aim of the following modification, \Cref{Alg.2}, is to overcome the bottleneck of MIG evaluation, we fixed the momentum factor $\mu = 1$ and baseline image $b = 0$.

\begin{algorithm}
	\caption{MIG modified version}\label{Alg.2}
	\begin{algorithmic}[1]
		\Require: model $f$ , perturbation budgets array $\epsilon$ with size $n$, number of iterations $T$, original image $x$, and its label $y$.
		\State create three arrays, each of size $n$, for step sizes $\alpha$, adversarial images $adv$, and gradients $g$.
        \For{$i = 1 \dots n$} \Comment{$g^{(i)}_t$ where $i$ is perturbation index, $t$ is iteration}
		\State $Adv^{(i)}_{0} = x$, $g^{(i)}_{0} = 0$, $\alpha^{(i)} = \frac{\epsilon^{(i)}}{T}$ 
		\EndFor
		\For{$t = 1 \dots T$} 
		\State $\Delta_t = IG(f,Adv^{(n)}_{t-1}, y)$  
		\For{$i = 1 \dots n$}
		\State $g^{(i)}_t = g^{(i)}_{t-1} + \frac{\Delta_t}{||\Delta_t||_1}$  
		\State $Adv^{(i)}_t = Clip_{\epsilon^{(i)}}(Adv^{(i)}_{t-1} + \alpha^{(i)}*sign(g^{(i)}_t))$ 
		\EndFor
		\EndFor
		\State Return $Adv$
	\end{algorithmic}
\end{algorithm}

\par
\begin{figure}[tbhp!]
	\centering
	\begin{multicols}{2}
		\includegraphics[width=\linewidth]{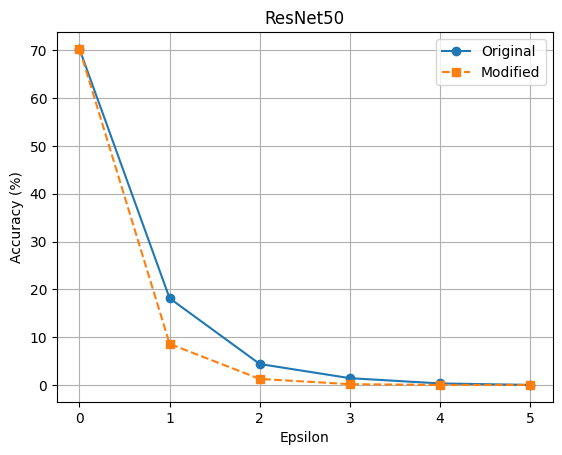}   \par 
		\includegraphics[width=\linewidth]{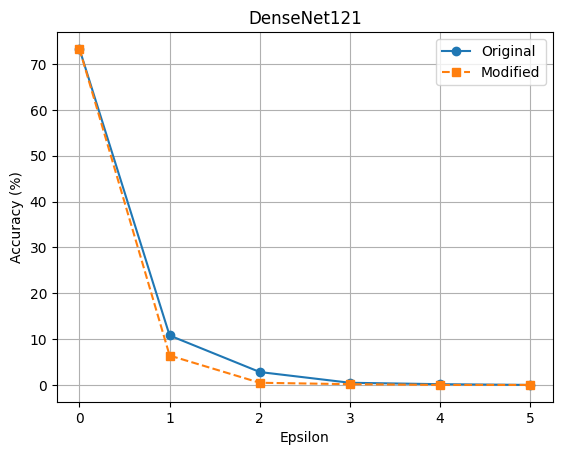}\par 
	\end{multicols}
    \vspace{-15pt}
	\caption{Top-1 Accuracy for original against modified MIG tested  using ResNet50 and DenseNet121.}
	\label{Figure 5.}
\end{figure}

 
The core idea is to compute the Integrated Gradients (IG) for the highest epsilon only, and then to perform the clipping step for each epsilon separately. 
This approach proved advantageous for our task, as it allowed us to generate adversarial images for five different epsilon values with a constant additional time. Empirical results demonstrated an increase in attack power at lower epsilon values compared to the original method, without compromising imperceptibility. We compared the original and modified algorithms on 640 images using the ResNet50 and DenseNet121 models, as illustrated in \Cref{Figure 5.}.

\appsection{Complete tables}
\label{complete tables}
\vspace{-10pt}
we present the full tables for reference. 

\begin{table*}[ht!]
	\centering
	\caption{Top-1 accuracy(\%) of the AT model on the clean dataset, against PGD-20 and MIG at different $\epsilon$.} 
	\label{table.2}
    \vspace{5pt}
	\tiny
	\begin{tabular}{| c | c | *{5}{c} | *{5}{c} | c |}
		\hline
		Model & Clean  & & & PGD-20 & & & & & MIG & & & Total \\
		\cline{3-12}
		Name & Adv. & $\epsilon = 1$ & $\epsilon = 2$ & $\epsilon = 3$ & $\epsilon = 4$ & $\epsilon = 5$ & $\epsilon = 1$ & $\epsilon = 2$ & $\epsilon = 3$ & $\epsilon = 4$ & $\epsilon = 5$ & Params \\
		\hline
		DenseNet121 & 63.51 & 48.01 & 33.33 & 21.54 & 13.30 & 8.14 & 50.13 & 38.12 & 27.63 & 20.75 & 16.63 & 8.1M \\
		DenseNet169  & 64.37 & 49.19 & 34.73 & 22.92 & 14.64 & 9.23 & 51.06 & 38.56 & 28.87 & 23.31 & 18.44 & 14.3M \\
		DenseNet201  & 64.89 & 49.89 & 35.32 & 23.83 & 15.41 & 9.71 & 52.69 & 40.38 & 30.25 & 24.19 & 20.00 & 20.2M \\
		\hline
		ResNet50  & 60.59 & 45.33 & 31.49 & 20.56 & 13.21 & 8.15 & 48.31 & 37.44 & 26.37 & 20.00 & 16.12 & 25.6M \\
		ResNet101 & 61.50 & 46.33 & 32.04 & 21.06 & 13.37 & 8.38 & 49.00 & 36.50 & 26.50 & 20.69 & 16.81 & 44.7M \\
		ResNet152  & 62.19 & 46.98 & 33.00 & 21.99 & 14.06 & 8.82 & 49.63 & 37.31 & 27.88 & 21.50 & 17.94 & 60.4M \\
		\hline
        ResNetRS50  & 67.36 & 53.02 & 38.71 & 27.09 & 18.18 & 11.89 & 56.44 & 44.81 & 35.37 & 28.75 & 23.69 & 35.7M \\
		ResNetRS101  & 68.59 & 54.26 & 40.01 & 28.13 & 18.97 & 12.70 & 58.31 & 45.63 & 36.94 & 29.56 & 24.88 & 63.7M \\
		ResNetRS152 & 67.58 & 53.48 & 39.65 & 27.90 & 19.00 & 12.54 & 58.06 & 45.44 & 36.00 & 28.94 & 24.12 & 86.8M \\
		\hline
		EfficientNetB0  & 64.66 & 49.83 & 34.82 & 22.47 & 13.59 & 7.94 & 51.56 & 38.00 & 28.06 & 19.94 & 16.06 & 5.3M \\
		EfficientNetB1  & 67.24 & 52.57 & 37.24 & 24.12 & 14.86 & 8.82 & 51.75 & 37.75 & 28.12 & 19.75 & 15.00 & 7.9M \\
		EfficientNetB2  & 66.56 & 52.20 & 37.02 & 24.36 & 14.91 & 8.82 & 54.37 & 39.94 & 29.50 & 22.88 & 17.94 & 9.2M \\
		EfficientNetB3  & 67.19 & 53.29 & 38.51 & 25.95 & 16.42 & 9.86 & 54.81 & 42.31 & 32.50 & 24.38 & 20.19 & 12.3M \\
		EfficientNetB4  & 68.07 & 53.98 & 39.15 & 26.47 & 16.67 & 10.12 & 55.44 & 42.50 & 32.56 & 26.44 & 21.25 & 19.5M \\
		EfficientNetB6 & 68.74 & 55.11  & 40.55 & 27.97 & 18.38 & 11.45 & 56.75 & 44.31 & 33.94 & 27.13 & 22.12 & 43.3M \\
		EfficientNetB7 & 68.60 & 55.18 & 40.84 & 28.02 & 18.54 & 11.84 & 57.69 & 44.00 & 34.38 & 27.81 & 23.75 & 66.7M \\
		\hline
        EfficientNetV2B0 & 66.10 & 51.17 & 36.36 & 23.71 & 14.61 & 8.49 & 53.19 & 40.62 & 30.81 & 22.62 & 18.44 & 7.2M \\
		EfficientNetV2B1 & 66.99 & 52.43 & 37.34 & 24.75 & 15.53 & 9.20 & 53.56 & 42.25 & 32.12 & 25.75 & 20.13 & 8.2M \\
		EfficientNetV2B2 & 68.55 & 53.64 & 37.89 & 24.97 & 15.34 & 9.06 & 56.19 & 42.56 & 32.62 & 25.56 & 20.94 & 10.2M \\
		EfficientNetV2B3 & 68.15 & 53.82 & 38.71 & 25.79 & 16.24 & 9.74 & 55.62 & 42.69 & 33.63 & 26.94 & 21.88 & 14.5M \\
		EfficientNetV2S & 68.11 & 54.15 & 39.76 & 27.33 & 17.85 & 11.25 & 56.19 & 44.37 & 33.56 & 26.25 & 21.63 & 21.6M \\
		EfficientNetV2M & 67.38 & 54.01 & 40.16 & 27.75 & 18.33 & 11.67 & 54.81 & 43.63 & 34.75 & 27.69 & 23.87 & 54.4M \\
		EfficientNetV2L & 66.53 & 52.16 & 37.70 & 25.91 & 16.97 & 10.71 & 54.87 & 41.31 & 32.44 & 26.75 & 22.50 & 119.0M \\
		\hline
		MobileNet & 59.99 & 44.05 &  29.58 & 18.53 & 11.19 & 6.43 & 45.06 & 32.88 & 24.06 & 17.00 & 14.06 & 4.3M \\
		MobileNetV2 & 58.52 & 43.50 & 29.97 & 19.44 & 11.67 & 6.87 & 45.75 & 34.13 & 24.75 & 17.69 & 14.75 & 3.5M \\
		MobileNetV3Small & 56.59 & 41.48 & 27.41 & 16.42 & 9.30 & 5.05 & 
		43.06 & 30.06 & 20.19 & 15.19 & 11.56 & 2.6M \\
		MobileNetV3Large & 58.81 & 43.89 & 29.83 & 18.68 & 11.19 & 6.30 & 45.19 & 32.19 & 23.69 & 18.31 & 14.19 & 5.5M \\
		\hline
		NASNetMobile & 56.70 & 42.41 & 29.47 & 19.01 & 11.90 & 7.25 & 42.88 & 31.37 & 23.00 & 18.50 & 14.31 & 5.3M \\
		NASNetLarge & 63.53 & 50.93 & 38.28 & 27.3 & 18.99 & 12.80 & 52.63 & 42.69 & 33.50 & 26.06 & 21.50 & 88.9M \\
		\hline
		ViT-Tiny/16 & 59.89 & 42.28 & 25.99 & 14.39 & 7.47 & 3.77 & 44.25 & 29.81 & 19.06 & 12.31 & 9.13 & 5.7M \\
		ViT-Tiny/32 & 50.1 & 33.98 & 20.17 & 10.88 & 5.54 & 2.79 & 36.13 & 22.88 & 14.44 & 9.31 & 6.25 & 6.1M \\
		ViT-Small/16 & 67.75 & 52.07 & 35.99 & 22.51 & 13.16 & 7.36 & 54.06 & 40.56 & 29.94 & 22.12 & 16.25 & 22.1M \\
		ViT-Small/32 & 58.28 & 41.56 & 26.44 & 15.47 & 8.42 & 4.47 & 54.06 & 40.56 & 29.94 & 22.12 & 16.25 & 22.9M \\
		ViT-Base/16 & 68.31 & 52.46 & 36.44 & 23.17 & 13.74 & 7.79 & 55.87 & 41.94 & 33.19 & 26.12 & 20.31 & 86.6M \\
		ViT-Base/32 & 59.25 & 42.55 & 27.51 & 16.46 & 9.16 & 4.96 & 45.12 & 32.75 & 24.06 & 17.5 & 13.81 & 88.2M \\
		ViT-Large/16 & 72.35 & 54.65 & 36.62 & 22.21 & 12.55 & 6.74 & 59.94 & 45.06 & 35.62 & 29.00 & 24.25 & 304.3M  \\
		\hline
	\end{tabular}
    \vspace{-15pt}
\end{table*}

\begin{table*}[t!]
	\centering
	\scriptsize
	\caption{Top-1 accuracy ($\%$) of large-capacity models on the transferability benchmark, presenting results for four modes of transfer attacks. These modes represent different combinations where the surrogate model is either normally trained or AT, and the target model is also either ST or AT. The table includes accuracy on both clean datasets and datasets generated by applying MIG with $\epsilon$ = 16 on the surrogate models.}
	\label{transfer table2}
    \centering
	\begin{tabular}{ c | c | c | *{4}{c} | *{4}{c} |}
		\cline{2-11}
		& Target Model & Clean &  & Surrogate & ST  &  &  & Surrogate & AT &  \\
		\cline{4-11}
		
		& Name & Dataset   & DN121 & ENB0 & MN & ViT-Ti/16 & DN121 & ENB0 & MN & ViT-Ti/16 \\
		\cline{2-11}
		& DN201 & 74.80 & \textbf{8.45} & 15.05 & 27.00 & 34.25 & \textbf{11.95} & 10.10 & 17.30 & 24.45 \\
		& RN152 & 72.80 & 17.30 & 16.35 & 28.85 & 34.60 & 11.35 & 9.20 & 14.95 & 22.65 \\
		\multirow{3}{*}{\rotatebox[origin=c]{90}{\centering ST}} & RN152RS & 80.60 & 30.65 & 23.75 & 43.80 & 43.55 & 25.80 & 20.30 & 30.15 & 37.60 \\
		& ENB7 & 76.20 & \textbf{26.00} & \textbf{14.35} & 37.70 & 41.80 & \textbf{27.35} & \textbf{22.90} & 33.10 & 38.25 \\
		& ENV2L & 84.35 & 46.85 & 39.45 & 60.35 & 59.40 & 42.55 & 38.90 & 48.90 & 58.15 \\
		& MNV3Large & 73.05 & 18.60 & \textbf{5.35} & 21.30 & 26.25 & 11.60 & \textbf{5.85} & 11.05 & 18.20 \\
		& NASNLarge & 80.60 & 27.15 & 24.20 & 40.85 & 47.85 & 23.60 & 20.35 & 29.80 & 40.65 \\
        & ViT-B/16 & 83.80 & 53.05 & 43.85 & 60.35 & 33.90 & 34.60 & 29.65 & 39.95 & 39.45 \\
		\cline{2-11}
		& DN201 & 64.10 & 46.60 & 43.50 & 46.95 & 48.70 & 6.85 & 7.35 & 8.95 & 25.05 \\
		& RN152 & 61.05 & 47.75 & 44.45 & 47.45 & 49.60 & 8.00 & 8.30 & 8.00 & 23.15 \\
		\multirow{4}{*}{\rotatebox[origin=c]{90}{\centering AT}} & RN152RS & 67.45 & 52.05 & 48.80 & 52.20 & 52.65 & 11.05 & 9.20 & 12.50 & 26.05 \\
		& ENB7 & 68.05 & 48.15 & 43.10 & 48.80 & 51.25 & 11.75 & 6.40 & 12.45 & 27.80 \\
		& ENV2L & 66.45 & 49.70 & 44.15 & 49.30 & 51.50 & 10.95 & 6.75 & 12.10 & 26.00 \\
		& MNV3Large & 57.15 & 44.60 & 38.85 & 43.45 & 42.85 & 9.80 & 5.00 & 6.85 & 15.35 \\
		& NASNLarge & 62.15 & 48.75 & 44.50 & 47.35 & 49.10 & 9.85 & 7.70 & 9.70 & 23.30 \\
        & ViT-B/16 & 68.15 & 54.50 & 53.10 & 57.00 & 48.30 & 21.50 & 16.80 & 19.60 & 6.90\\
		\cline{2-11}
	\end{tabular}
\end{table*}


\end{document}